# Texture Defect Detection in Gradient Space

Dr. V. Asha
Dept. of Computer Applications
New Horizon College of Engineering
Bangalore, Karnataka, INDIA
v_asha@live.com

Dr. N.U. Bhajantri
Dept. of Computer Science and Engineering
Govt. College of Engineering, Chamarajanagar
Mysore District, Karnataka, INDIA
bhjan3nu@gmail.com

Dr. P. Nagabhushan
Dept. of Studies in Computer Science
University of Mysore
Mysore, Karnataka, INDIA
pnagabhushan@hotmail.com

*Abstract—* **In this paper, we propose a machine vision algorithm for automatically detecting defects in patterned textures with the help of gradient space and its energy. Gradient space image is obtained from the input defective image and is split into several blocks of size same as that of the periodic unit of the input defective image. Energy of the gradient space image is used as feature space for identifying defective and nondefective periodic blocks using Ward's hierarchical clustering. Experiments on real fabric images with defects show that the proposed method can be used for automatic detection of fabric defects in textile industries.**

## I. INTRODUCTION

Product inspection is a major concern in quality control of various industrial products such as textile fabrics and ceramic tiles. Among various industries, textile industry is one of the biggest traditional industries requiring automated inspection system which not only increases the production efficiency by decreasing the inspection time but also reduces the storage space of the reference images needed for cross-referencing.

Though human-vision based inspection is common, lack of reproducibility of inspection results due to fatigue and subjective nature of human inspections, prolonged inspection time and lack of perfect defect detection due to complicated design in textile patterns are certain drawbacks in human-vision based inspection. Due to complexity in the design and existence of numerous categories of patterns in the textile fabrics (such as patterned fabrics as shown in Fig. 1) produced by modern textile industries and similarity between the defect and background, inspection on patterned textures has become more complicated [1]. So, most of the methods in literature rely on training stage with numerous defect-free samples for obtaining decision-boundaries or thresholds [1-6].

In this paper, we make use of gradient space domain of the input texture image to discern between defective and nondefective patterns. Gradient space of an image gives a measure of change in intensity over the pixels rather than the absolute values of the pixel intensities [7]. A break in the regularity of image of a patterned texture will have more enhanced effect in gradient space domain than in image space domain. As a result, feature extracted from gradient space of a defective periodic block will have significant difference with reference to that of a nondefective block. Considering this fact, gradient space domain is utilized in the proposed algorithm for finding defects in images of patterned textures. The program for the proposed algorithm is written in Matlab-7.0.1 and run in a Pentium-4 personal computer of 2 GB RAM capacity and 1.8 GHz clock speed.

The organization of the paper is as follows: Section-II presents a brief review on gradient space domain. Section-III presents the proposed method of defect detection, illustration of the proposed method, results from experiments on various real fabric images with defects and the performance evaluation of the proposed algorithm for defect detection. Section-IV has the conclusions.

## II. GRADIENT SPACE DOMAIN

Gradient of an image is the measure of its edge strength. For an image function $f(x, y)$, the gradient of $f$ at $(x, y)$ is a two-dimensional vector given as [7]

$$\nabla f = \mathrm{grad}(f) = \begin{pmatrix} g_x \\ g_y \end{pmatrix} = \begin{pmatrix} \dfrac{\partial f}{\partial x} \\ \dfrac{\partial f}{\partial y} \end{pmatrix}. \qquad (1)$$

This vector is a geometrical property that it points in the direction of the gradient vector. The magnitude of the vector is referred to as *gradient space* and is given by

$$|\nabla f| = \sqrt{g_x^2 + g_y^2}. \qquad (2)$$

The partial derivatives required for calculating the gradient at any pixel location $(x, y)$ can be obtained by discretizing the partial derivatives along the two orthogonal directions.

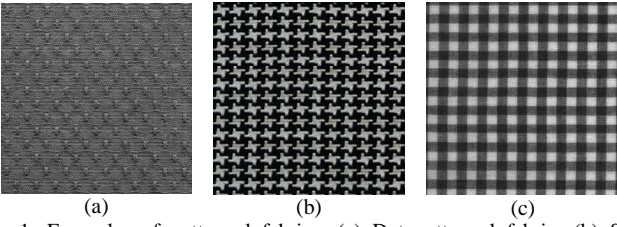

Fig. 1. Examples of patterned fabrics: (a) Dot patterned fabric; (b) Star patterned fabric; (c) Box patterned fabric.

## III. PROPOSED DEFECT DETECTION ALGORITHM

### A. Description of the algorithm

There are three main assumptions in the proposed algorithm as follows:
(i) Test image is of at least two periodic units in horizontal direction and two in vertical direction whose dimensions are known apriori.
(ii) Number of defective periodic units is always less than the number of defect-free periodic units.
(iii) Test images are from imaging system oriented perpendicular to the surface of the product such as textile fabric. This assumption is due to the fact that in a defect detection system in industries such as fabric industry, the imaging system is always oriented perpendicular to the plane of fabric surface.

An image under inspection may have fractional periodic units also. Neglecting the fractional portions will result in missing of defective portions, if any, in the fractional portions. So, four cropped images are obtained from the resultant image to get complete periodic units and each cropped image is analyzed for identifying defect based on [8]. If $g$ is the resultant gradient space domain of an image of size $M \times N$ with row periodicity $P_r$ (i.e., number of columns in a periodic unit) and column periodicity $P_c$ (i.e., number of rows in a periodic unit), then the size of cropped image $g_{crop}$ is $M_{crop} \times N_{crop}$ where $M_{crop}$ and $N_{crop}$ are measured from all four corners (top-left, bottom-left, top-right and bottom-right) and are given by the following equations:

$$M_{crop} = \text{floor}(M / P_c) \times P_c. \quad (3)$$

$$N_{crop} = \text{floor}(N / P_r) \times P_r. \quad (4)$$

Each cropped image is split into several periodic blocks of size $P_c \times P_r$. Energy of each block in L1 norm [9] is used as a feature space for Ward's hierarchical clustering which is based on inner squared distance and minimum variance criterion [10]. The cluster algorithm yields a cluster tree in the form of a linkage matrix $Z$ of size $(n - 1) \times 3$, where $n$ is the total number of periodic blocks. The leaf nodes in the cluster hierarchy are the periodic blocks in the original data set numbered from 1 to $n$. These are the singleton clusters from which all higher clusters are built. Each newly formed cluster, corresponding to row $i$ in $Z$, is assigned the index $n + i$. The first and second columns of the linkage matrix contain the indices of the periodic blocks that were linked in pairs to form a new cluster. This new cluster is assigned the index value $n + i$. There are $n - 1$ higher clusters corresponding to the interior nodes of the hierarchical cluster tree. The third column contains the corresponding linkage distances between the periodic blocks paired in the clusters at each row $i$. The last value of the linkage distance is maximum indicating that all periodic blocks are grouped into one cluster and the last but one value of the linkage distance is the next maximum corresponding to two clusters yielding the required cut-off point at which two clusters (one cluster containing defective periodic blocks and other cluster containing nondefect periodic blocks) are formed. Upon identifying the two clusters, the number of periodic blocks in one cluster is compared with that of other and the cluster with less number of periodic blocks is assumed to be defective. Detection of defective blocks from each cropped image does not give an overview of the total defects in the input defective image. Hence, in order to get the overview of the total defects in the input image, we use the concept of *fusion* of defects proposed in [8] that involves merging of boundaries of the defective blocks identified from each cropped image, morphological filling [7] and Canny edge detection [7]. The entire procedure for defect detection is shown in the form of flow chart in Fig. 2.

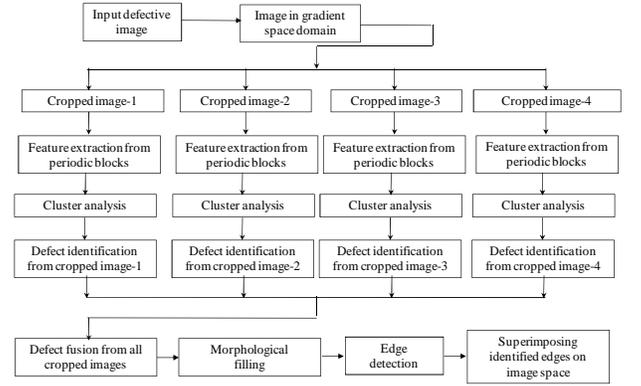

Fig. 2. Proposed inspection scheme.

### B. Illustration of the algorithm

In order to illustrate the proposed algorithm, let us a defective dot-patterned fabric image shown in Fig. 3 (a). The partial derivatives required for calculating the gradient of the image at any pixel location $(x, y)$ are calculated using Newton's forward difference scheme as follows:

$$g_x = \frac{\partial f(x, y)}{\partial x} = f(x+1, y) - f(x, y). \quad (5)$$

$$g_y = \frac{\partial f(x, y)}{\partial y} = f(x, y+1) - f(x, y). \quad (6)$$

Fig. 3 (b) shows the magnitude of the gradient for the defective image thus calculated using masks $(-1,1)$ and $(-1,1)^T$ for horizontal and vertical directions respectively. Following (3) an (4), four cropped images containing complete number of periodic blocks are obtained from the gradient space of the original image with the help of periodicities known apriori. Each cropped image is split into several blocks of size same as the size of the periodic unit. Fig. 4 shows the cropped images highlighting boundaries of the defective blocks identified using energy calculated in L1 norm of each periodic block as feature space for Ward's clustering. The boundaries of defective periodic blocks identified from each cropped image are superimposed on the gradient space as shown in Fig. 5 (a) and separately on plain background as shown in Fig. 5 (b). These zones are morphologically filled (Fig. 5 (c)) and their edges are extracted using Canny's edge operator as shown in Fig. 5 (d). The extracted edges are shown superimposed on gradient space and on original defective image as shown in Fig. 5 (e) and Fig. 5 (f) respectively. Fig. 6 shows additional results from test images of defective dot-patterned, star-patterned and box-patterned fabrics.

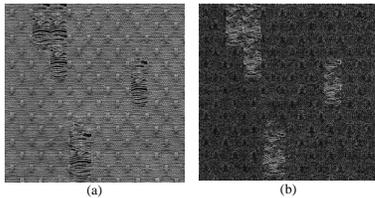

Fig. 3: (a) Defective dot patterned fabric; (b) Its gradient space.

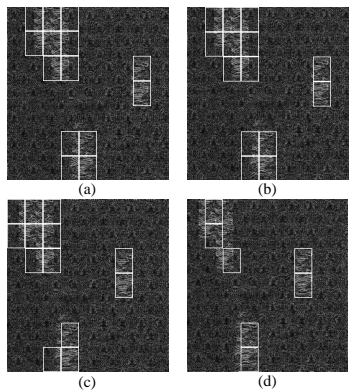

Fig. 4. Defective periodic blocks identified from cropped image obtained from (a) top left corner, (b) bottom left corner, (c) top right corner and (d) bottom right corner of the gradient space of the defective dot patterned image.

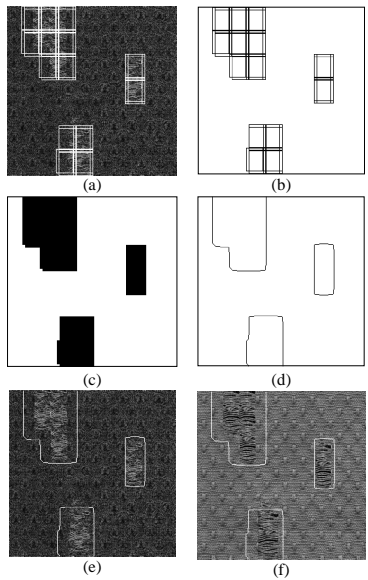

Fig. 5. Illustration of defect fusion: (a) Boundaries of the defective blocks identified from each cropped image shown superimposed on the original gradient space image; (b) Boundaries of the defective blocks shown separately on plain background; (c) Result of morphological filling; (d) Canny edge detection; (e) Identified edges shown superimposed on gradient space; (f) Identified edges shown superimposed on original defective image.

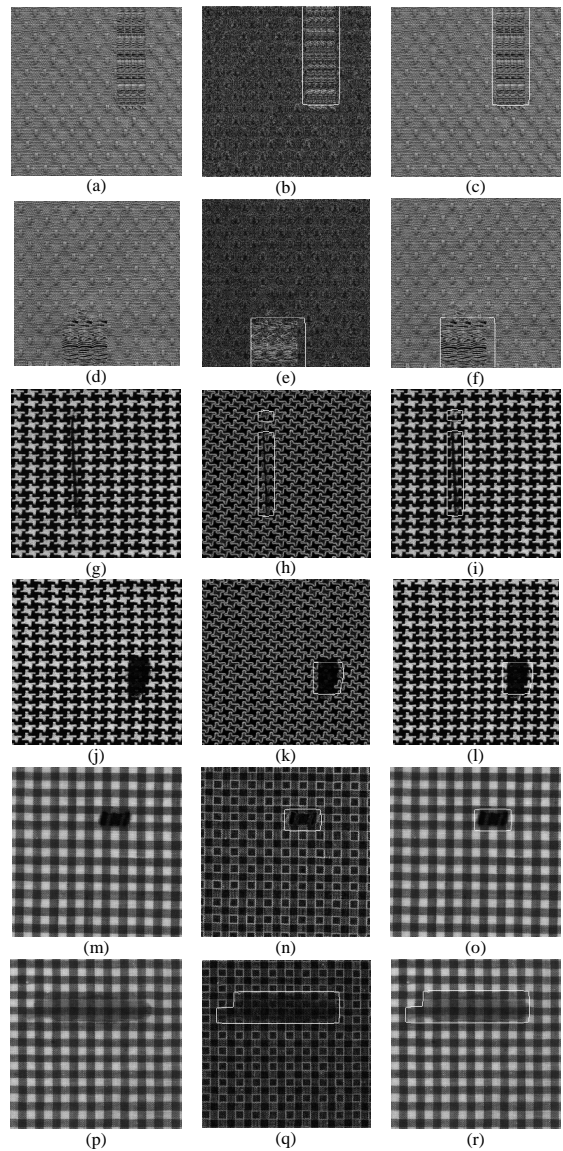

Fig. 6. Result of defect detection on few real fabric images: Left column shows the defective test images; Middle column shows the final result of defect detection in gradient space domain; Right column shows the edges of the defects superimposed on original defective image; (a) and (d) represent defective dot patterned fabrics; (g) and (j) represent defective star patterned fabrics; (m) and (p) represent defective star patterned fabrics.

*C. Performance evaluation of the proposed algorithm*

Precision, recall and accuracy are commonly used performance measures in several retrieval applications [11]. In order to access the performance of the proposed method of defect detection, these performance parameters are calculated in terms of true positive (TP), true negative (TN), false positive (FP), and false negative (FN), where true positive is the number of defective periodic blocks identified as defective, true negative is the number of defect-free periodic blocks identified as defect-free, false positive is the number of defect-free periodic blocks identified as defective and false negative is the number of defective periodic blocks identified as defect-free. Precision is defined as the number of periodic blocks correctly labeled as belonging to the positive class divided by the total number of periodic blocks labeled as belonging to the positive class and is calculated as TP/(TP+FP). Recall is defined as the number of true positives divided by the sum of true positives and false negatives that are periodic blocks not labeled as belonging to the positive class but should have been and is calculated as TP/(TP+FN). Accuracy is the measure of correctness of detection and is calculated as (TP+TN)/(TP+TN+FP+FN). Though the number of periodic blocks taken from a defective input image is same for all of its cropped images, the number of defective periodic blocks identified does not need to be same for all cropped images. This is because the contribution of defect in each periodic block may differ for different cropped images. The performance parameters averaged over all cropped images for each defective image are given in Table 1. The average precision, recall and accuracy rates based on 2136 periodic blocks during testing stage are 100%, 85.9% and 97.6% respectively for all the test images. Relatively less recall rate indicates that there are few false rates in the results. However, high precision and accuracy indicate that the proposed algorithm can contribute to automated inspection scheme in fabric industries.

TABLE I. PERFORMANCE PARAMETERS FOR EACH TEST IMAGE

| Test image | No. of periodic blocks | Precision (%) | Recall (%) | Accuracy (%) |
|---|---|---|---|---|
| Fig. 3 (a) | 252 | 100 | 80.0 | 96.8 |
| Fig. 6 (a) | 252 | 100 | 73.8 | 93.7 |
| Fig. 6 (d) | 252 | 100 | 100 | 100 |
| Fig. 6 (g) | 330 | 100 | 90.0 | 99.4 |
| Fig. 6 (j) | 330 | 100 | 78.6 | 99.1 |
| Fig. 6 (m) | 360 | 100 | 78.9 | 94.4 |
| Fig. 6 (p) | 360 | 100 | 100 | 100 |

IV. CONCLUSIONS

In this paper, we have proved that gradient space of an image and its energy can be effectively used for finding defects in patterned textures. Results of experiments on real fabric images with defects show the capability of the proposed algorithm for defect detection. High precision and accuracy of the proposed algorithm for defect detection without any training stage indicate that the proposed method can contribute to automated defect detection scheme in fabric industries.

ACKNOWLEDGMENT

The authors would like to thank Dr. Henry Y. T. Ngan, Research Associate of Industrial Automation Research Laboratory, Department of Electrical and Electronic Engineering, The University of Hong Kong, for providing the database of patterned fabrics.

REFERENCES

[1] H. Y. T. Ngan and G. K. H. Pang, "Regularity Analysis for Patterned Texture Inspection," *IEEE Trans. on Autom. Sci. Eng.*, vol. 6, no. 1, pp. 131-144, 2009.
[2] H. Y. T. Ngan and G. K. H. Pang, "Novel method for patterned fabric inspection using Bollinger bands," *Opt. Eng.*, vol. 45, no. 8, Aug. 2006.
[3] F. Tajeripour, E. Kabir, and A. Sheikhi, "Fabric Defect Detection Using Modified Local Binary Patterns," in *Proc. of the Int. Conf. on Comput. Intel. Multiméd. Appl.*, vol. 2, Dec. 2007, pp. 261-267.
[4] H. Y. T. Ngan, G. K. H. Pang and N.H.C. Yung, "Motif-based defect detection for patterned fabric," *Pattern Recognit.*, vol. 41, pp. 1878-1894, 2008.
[5] H. Y. T. Ngan, G. K. H. Pang and N. H. C. Yung, "Ellipsoidal decision regions for motif-based patterned fabric defect detection," *Pattern Recognit.*, vol. 43, pp. 2132-2144, 2010.
[6] H. Y. T. Ngan, G. K. H. Pang and N. H. C. Yung, "Performance Evaluation for Motif-Based Patterned Texture Defect Detection," *IEEE Trans. on Autom. Sci. Eng.*, vol. 7, no. 1, pp. 58-72, 2010.
[7] R. C. Gonzalez and R. E. Woods, *Digital Image Processing*, Third Edition, Pearson Prentice Hall, New Delhi, 2008.
[8] V. Asha, N. U. Bhajantri and P. Nagabhushan, "Automatic Detection of Texture Defects using Texture-Periodicity and Gabor Wavelets," in: K.R. Venugopal and L.M. Patnaik (Eds.): ICIP 2011, *Communication and Computer Information Series* (CCIS), vol. 157, Springer-Verlag, Berlin Heidelberg, pp. 548–553, 2011.
[9] Wikipedia, http://en.wikipedia.org/wiki/Lp_space, 2010.
[10] Wikipedia, http://en.wikipedia.org/wiki/Cluster_analysis, 2010.
[11] Wikipedia, http://en.wikipedia.org/wiki/Precision_and_recall, 2010.